\documentclass[conference]{IEEEtran}
\IEEEoverridecommandlockouts

\usepackage{cite}
\usepackage{amsmath,amssymb,amsfonts}
\usepackage{graphicx}
\usepackage{textcomp}
\usepackage{xcolor}

\usepackage{kotex}
\usepackage{inconsolata}
\definecolor{vscodegreen}{RGB}{0,110,0}
\newcommand{\comment}[1]{\textcolor{vscodegreen}{\ttfamily\footnotesize #1}}

\usepackage{hhline}
\usepackage{subcaption}
\usepackage{amsmath}
\usepackage{amssymb}
\usepackage{mathtools}
\usepackage{placeins}
\usepackage{float}
\usepackage{algorithm}
\usepackage{circledsteps}
\pgfkeys{
  /csteps/inner color=white,
  /csteps/outer color=black,
  /csteps/fill color=black
}
\usepackage{setspace}
\usepackage{algpseudocode}
\algtext*{EndFor}
\algtext*{EndIf}

\def\BibTeX{{\rm B\kern-.05em{\sc i\kern-.025em b}\kern-.08em
    T\kern-.1667em\lower.7ex\hbox{E}\kern-.125emX}}
\begin{document}

\title{Training Time Prediction for Mixed Precision-based Distributed Training
}

% \and
% \and
% \and
% \and
% \and

\author{\parbox{\textwidth}{\centering
\begin{minipage}[t]{0.23\textwidth}\centering
Minchul Kang\\
\textit{Korea University}\\
mckang@os.korea.ac.kr
\end{minipage}\hspace{0.02\textwidth}
\begin{minipage}[t]{0.23\textwidth}\centering
Changyong Shin\\
\textit{Korea University}\\
cyshin@os.korea.ac.kr
\end{minipage}\hspace{0.02\textwidth}
\begin{minipage}[t]{0.23\textwidth}\centering
Jinwoo Jeong\\
\textit{Korea University}\\
jwjeong@os.korea.ac.kr
\end{minipage}\hspace{0.02\textwidth}
\begin{minipage}[t]{0.23\textwidth}\centering
Hyunho Lee\\
\textit{Korea University}\\
hhlee@os.korea.ac.kr
\end{minipage}

\vspace{0.8em}

\begin{minipage}[t]{0.23\textwidth}\centering
Younghun Go\\
\textit{Korea University}\\
yhgo@os.korea.ac.kr
\end{minipage}\hspace{0.02\textwidth}
\begin{minipage}[t]{0.23\textwidth}\centering
Gyeongmin Kim\\
\textit{KT Corporation}\\
gyeongmin.kim@kt.com
\end{minipage}\hspace{0.02\textwidth}
\begin{minipage}[t]{0.23\textwidth}\centering
Gyeongsik Yang\\
\textit{Korea University}\\
g\_yang@korea.ac.kr
\end{minipage}\hspace{0.02\textwidth}
\begin{minipage}[t]{0.23\textwidth}\centering
Chuck Yoo\\
\textit{Korea University}\\
chuckyoo@os.korea.ac.kr
\end{minipage}
}}

\maketitle

\begin{abstract}
Accurate prediction of training time in distributed deep learning is crucial for resource allocation, cost estimation, and job scheduling. We observe that the floating-point precision setting is a key determinant of training time, leading to training time variations of $\sim$2.4$\times$ over its minimum. However, existing studies on distributed training time prediction rely on static model computation graphs that do not capture precision variations, including mixed precision. According to our experiments, training time prediction without considering precision results in significant prediction errors---reaching up to 147.85\% in mean absolute percentage error (MAPE). To address this issue, we propose a precision-aware distributed training time predictor that achieves robust accuracy across diverse precision settings, including mixed precision, with 9.8\% MAPE.
\end{abstract}

% \begin{IEEEkeywords}
% component, formatting, style, styling, insert.
% \end{IEEEkeywords}

\section{Introduction}
As deep learning (DL) models continue to grow in scale, training large-scale models on a single GPU becomes infeasible. For instance, training GPT-3 175B on a single NVIDIA A100 GPU would take about 14.8 years \cite{intro}. 
Consequently, distributed training leveraging multiple GPUs has emerged as the de facto standard. Accurate prediction of training time\footnote{The training time in this work refers to single iteration time.} is essential for hardware resource planning \cite{driple}, efficient job scheduling \cite{tensorshare, lazer}, and cost estimation.

We observe that training time increases by $\sim$2.4$\times$ depending on the floating-point precision setting (e.g., FP32, FP16, and mixed precision), implying that precision must be considered for accurate prediction. However, state-of-the-art training time prediction studies assume a fixed precision setting \cite{neusight,vtrain}, leading to significant prediction errors (MAPE) of $\sim$147.85\%.

To this end, we propose a distributed training time predictor that supports arbitrary precision settings. By partitioning the model computation graph, the predictor automatically identifies operator-level precision, and incorporates communication overheads into the prediction. Our predictor achieves high accuracy, with 9.8\% MAPE across various precision settings.

\section{Background \& Motivation}
\subsection{Background}

\textbf{Parallelism strategies.} 
Modern distributed training commonly employs a combination of data parallelism (DP), tensor parallelism (TP), and pipeline parallelism (PP). In DP \cite{pytorch-distributed}, each GPU holds a full replica of the model and processes different input batches. TP \cite{megatron-lm-1} partitions the input dimensions of tensor operators within each model layer across multiple GPUs. PP \cite{gpipe} divides the model into sequential stages, each assigned to a different GPU, and processes micro-batches in a pipelined manner. The combination of these parallelisms significantly affects computation and communication time, as communication is known to be a bottleneck in distributed deep learning \cite{valo}.

\textbf{Floating-point precision.}
DL training uses various floating-point formats---FP32 for higher numerical accuracy and FP16 for faster computation. To balance FP32’s accuracy with FP16’s speed, mixed precision \cite{mixed-precision} is widely adopted, and recent large language models such as LLaMA \cite{llama-mixed}, Qwen \cite{qwen}, and GPT \cite{gpt} are trained under mixed precision. Mixed precision casts lower precision to compute-intensive operators (e.g., \texttt{conv}, \texttt{matmul}) and higher precision to operators requiring high numerical precision (e.g., \texttt{softmax}, \texttt{reduction}). 

\begin{figure}[t]
    \centering
    \includegraphics[width=\linewidth]{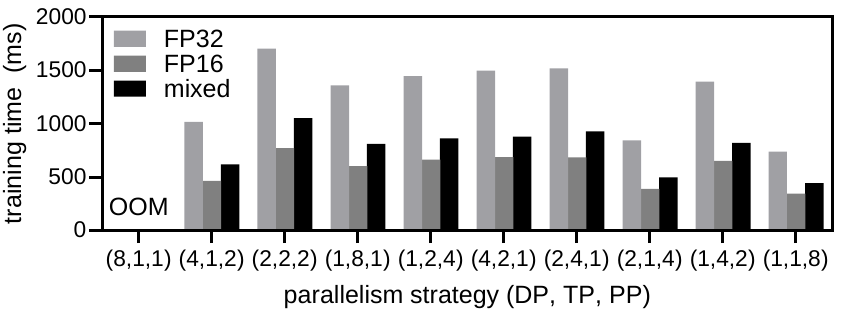}
    \caption{Training time (ms) by precision settings. OOM denotes a GPU out-of-memory error.}
    \label{fig:background}
\end{figure}

Fig.~\ref{fig:background} shows the training time of LLaMA~3.1--8B measured on eight NVIDIA H100 GPUs \cite{nvidia-h100} interconnected with NVLink \cite{nvlink}, under different parallelism strategies and floating-point precision settings. Training time varies by $\sim$2.4$\times$ across precision settings, highlighting its strong sensitivity to floating-point precision. This implies that accurate training-time prediction must account for the impact of floating-point precision.

\subsection{Motivation}
Prior work has proposed methods to predict the training time of DT. However, these methods rely heavily on static model computation graphs that do not capture precision variations such as mixed precision, which limits their applicability to general DT.

\begin{figure}[t]
    \centering
    \begin{subfigure}[t]{0.49\linewidth} 
        \centering
        \includegraphics[width=\linewidth]{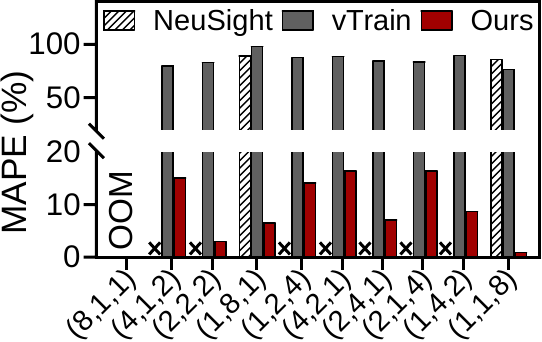}
        \caption{mixed precision.}
        \label{fig:3}
    \end{subfigure}
    \hfill
    \begin{subfigure}[t]{0.49\linewidth}
        \centering
        \includegraphics[width=\linewidth]{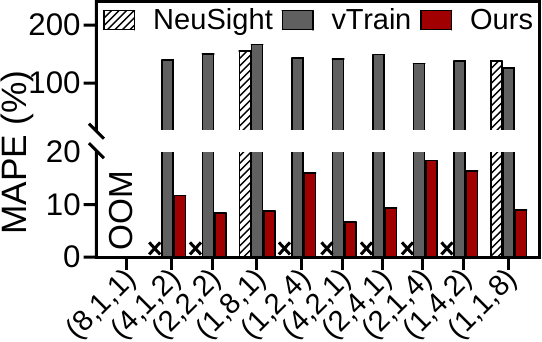}
        \caption{FP16.}
        \label{fig:4}
    \end{subfigure}
    \caption{Prediction error (MAPE) of existing works and proposed approach. (X, Y, Z) on the x-axis represent the degrees of DP, TP, and PP.}
    \label{fig:motivation}
\end{figure}

NeuSight \cite{neusight} decomposes the model computation graph of DL model into tiles, infers per-tile latency, and adds communication time for communication overheads. It assumes a fixed precision and, moreover, supports only a single parallelism strategy, not mixed parallelism (i.e., combinations of DP, TP, and PP). vTrain \cite{vtrain} converts the model computation graph to a CUDA graph, benchmarks kernel latencies, and incorporates communication overheads. However, like NeuSight, its prediction is based on a precision (FP32) and cannot cope with mixed or unseen precisions. We implement the prediction models of NeuSight and vTrain with FP32, and compare their predicted training times with the actual times observed under mixed precision and FP16 settings (Fig. \ref{fig:3}). The results demonstrate that both NeuSight and vTrain exhibit poor generalization to mixed and unseen precisions (i.e., FP16). Specifically, NeuSight's prediction error increases $\sim$130.55\%, while vTrain's error reaches $\sim$147.85\%. For unseen precisions, including mixed precision, input feature adjustments or reconstruction of the prediction model is inevitable.

\section{Design}

Our approach is to predict the execution time of the model computation graph and communication overheads introduced by the precision setting and parallelism strategies (DP, TP, and PP) as formulated in Eq.~\ref{eq:estimated_time}. Here, $T_{comp}$ denotes the computation graph execution time of the given model, and $T_{dp}$, $T_{tp}$, and $T_{pp}$ denote the communication overheads of DP, TP, and PP, respectively.

\begin{equation}
    T(d, t, p) = T_{comp}(d, t, p) + T_{dp}(d) + T_{tp}(t) + T_{pp}(p)
    \label{eq:estimated_time}
\end{equation}

\textbf{Computation graph execution time.} 
Given a model and a \textit{job config} that specifies the precision setting, parallelism strategies, and hyperparameters (e.g., batch size), we first extract unique operators such as \texttt{matmul}, \texttt{softmax} from the model using \texttt{torch.fx} library \cite{pytorch-fx}. Then, since the model computation graph can be distributed across multiple GPUs according to the given combination of parallelization strategies, we partition the graph into GPU-specific subgraphs based on the given DP, TP, and PP configurations, as detailed in Algorithm~\ref{algo:partition-subgraphs}.

Subsequently, we examine the precision of each operator within the subgraphs. If the training is configured for mixed precision, we hook the \texttt{torch.amp} \cite{pytorch-amp} library to determine the casted precision of each operator. Otherwise, if the precision is fixed to FP32 or FP16, we simply use that predefined precision. 
Since the tensor shapes of operators vary depending on hyperparameters, such as batch size, we profile each operator's execution time across both forward and backward computation graphs using the specific settings from the \textit{job config}. Finally, we aggregate the execution time of operators within each subgraph to obtain $T_{comp}$.

\begin{algorithm}[H]
\caption{Partition model into GPU-specific subgraphs}
\label{algo:partition-subgraphs}
\setstretch{0.9}
\begin{algorithmic}[1]
\State \textbf{Input:} Model $M$, Job config $J$
\State \textbf{Output:} $S$ -- set of GPU-specific subgraphs

\State \comment{// Initialize subgraph set and layers per pipeline stage}
\State $S \gets \emptyset$; $Lps \gets \lfloor layers(M)/J.PP \rfloor$

% \State \texttt{// Iterate over pipeline stages}
\For{$s=0$ \textbf{to} $J.PP-1$} \comment{// Iterate over pipeline stages}
    \State \comment{// Assign layers for stage $s$}
    \State $L_s \gets assign\_layers(M, s, Lps, J.PP)$
    
    \For{$d=0$ \textbf{to} $J.DP-1$}
        \For{$t=0$ \textbf{to} $J.TP-1$} \comment{// Create TP partition}
            \State $L'_s \gets copy(L_s)$
            
            \State \comment{// Slice weights if TP > 1}
            \ForAll{layer $u$ in $L'_s$}
                \ForAll{weight $w$ in $u$}
                    \If{$J.TP>1$ \textbf{and} SliceNeeded($u,w$)}
                        \State $L'_s\ \gets slice\_weight(w, t, J.TP)$ \label{line:slice_weight}
                    \EndIf
                \EndFor
            \EndFor
            \State $S \gets S \cup \{L'_s\}$
        \EndFor
    \EndFor
\EndFor
\State \textbf{return} $S$
\end{algorithmic}
\end{algorithm}

\textbf{Communication overhead.} 
In DP and TP, gradients are synchronized via all-reduce \cite{all-reduce}. Accordingly, we derive $T_{dp}$ and $T_{tp}$ as the communication volume (i.e., total gradient size) divided by the link bandwidth $B_{link}$ specified in the \textit{job config} as formulated in Eq.~\ref{eq:dp} and Eq.~\ref{eq:tp}. To determine $V_{dp}$, we aggregate the gradient sizes of all trainable parameters within the backward computation graph. Each size is precisely determined by the operator-level precision identified during the $T_{comp}$ prediction stage.  

\begin{equation}
    T_{dp} = V_{dp} / B_{link}
    \label{eq:dp}
\end{equation}

In contrast, $V_{tp}$ is determined by calculating the partial gradient sizes of operators partitioned in line~\ref{line:slice_weight} of Algorithm~\ref{algo:partition-subgraphs}.

\begin{equation}
    T_{tp} = V_{tp} / B_{link}
    \label{eq:tp}
\end{equation}

The main communication overhead of PP ($T_{pp}$) stems from the pipeline bubble, where stages wait for preceding computations to complete. We predict this overhead by scaling the predicted $T_{comp}$ by the PP degree specified in the \textit{job config}, as formulated in Eq.~\ref{eq:pp}. 

\begin{equation}
    T_{pp} = T_{comp} \times (PP-1)
    \label{eq:pp}
\end{equation}

\section{Evaluation}
We evaluate our predictor on the LLaMA 3.1-8B with C4 \cite{c4} dataset using eight NVIDIA H100 GPUs across all possible combinations of precision settings and parallelism strategies. In Fig.~\ref{fig:motivation}, the average MAPE values are 9.8\% for mixed precision and 10.64\% for unseen precision (FP16). Compared with existing approaches, our method achieves an $\sim$15.08$\times$ improvement, indicating robust and generalizable accuracy across diverse DT configurations.

\section{Conclusion \& Future Work}
We propose a distributed training time predictor that accurately predicts training time for arbitrary precision settings. By partitioning the model computation graph and profiling operator execution with precision applied to the operator, our predictor achieves 9.8\% average MAPE. 
% We plan to extend this study to predict training time in multi-node heterogeneous GPU environments.
We plan to extend this study to predict training time in multi-node heterogeneous GPU environments.

\section{Acknowledgment}
This work was supported by KT(Korea Telecom)-Korea University AICT R\&D Center.

\bibliographystyle{IEEEtran}
\bibliography{references}

\end{document}